\documentclass[10pt,twocolumn,letterpaper]{article}

\usepackage{iccv}
\usepackage{times}
\usepackage{epsfig}
\usepackage{graphicx}
\usepackage{amsmath}
\usepackage{amssymb}
\usepackage[table,xcdraw]{xcolor}
\usepackage{multirow}
\usepackage{tabularx} 
\usepackage{enumitem}
\usepackage{color}
\usepackage{graphics}
\usepackage{caption}
\usepackage{xspace}
\usepackage{subfigure}
\usepackage{bm} 
\usepackage{indentfirst} 
\usepackage{booktabs}
\usepackage{threeparttable}
\usepackage[accsupp]{axessibility}

\newcommand\blfootnote[1]{%
  \begingroup
  \renewcommand\thefootnote{}\footnote{#1}%
  \addtocounter{footnote}{-1}%
  \endgroup
}

\usepackage[pagebackref=true,breaklinks=true,letterpaper=true,colorlinks,bookmarks=false]{hyperref}

\iccvfinalcopy 

\def\iccvPaperID{4910} 

\ificcvfinal\fi

\begin{document}

\title{Towards Nonlinear-Motion-Aware and Occlusion-Robust \\ Rolling Shutter Correction}


\author{Delin Qu$^{1,2,\ast}$\quad Yizhen Lao$^{4,\ast}$\quad Zhigang Wang$^{2,\ast}$ \quad Dong Wang$^{2}$\quad Bin Zhao$^{2,3,\dagger}$ \quad Xuelong Li$^{2,3,\dagger}$\\
$^{1}$Fudan University\qquad$^{2}$Shanghai AI Laboratory\qquad \\ $^{3}$Northwestern Polytechnical University\qquad$^{4}$Hunan University
}

\maketitle
\ificcvfinal\thispagestyle{empty}\fi

\begin{abstract}
This paper addresses the problem of rolling shutter correction in complex nonlinear and dynamic scenes with extreme occlusion. Existing methods suffer from two main drawbacks. Firstly, they face challenges in estimating the accurate correction field due to the uniform velocity assumption, leading to significant image correction errors under complex motion. Secondly, the drastic occlusion in dynamic scenes prevents current solutions from achieving better image quality because of the inherent difficulties in aligning and aggregating multiple frames. To tackle these challenges, we model the curvilinear trajectory of pixels analytically and propose a geometry-based \textbf{Q}uadratic \textbf{R}olling \textbf{S}hutter (QRS) motion solver, which precisely estimates the high-order correction field of individual pixels. Besides, to reconstruct high-quality occlusion frames in dynamic scenes, we present a 3D video architecture that effectively \textbf{A}ligns and \textbf{A}ggregates multi-frame context, namely, RSA$^2$-Net. We evaluate our method across a broad range of cameras and video sequences, demonstrating its significant superiority. Specifically, our method surpasses the state-of-the-art by \textbf{+4.98}, \textbf{+0.77}, and \textbf{+4.33} of PSNR on Carla-RS, Fastec-RS, and BS-RSC datasets, respectively. Code is available at \href{https://github.com/DelinQu/QRSC}{https://github.com/DelinQu/qrsc}.

\blfootnote{$\ast$ Authors contributed equally: \href{mailto:dlqu22@m.fudan.edu.cn}{dlqu22@m.fudan.edu.cn}}
\blfootnote{$\dagger$ Corresponding author}
\end{abstract}

\section{Introduction}
\label{sec:introduction}
\begin{figure}[t]
    \begin{center}
        \includegraphics[width=1\linewidth]{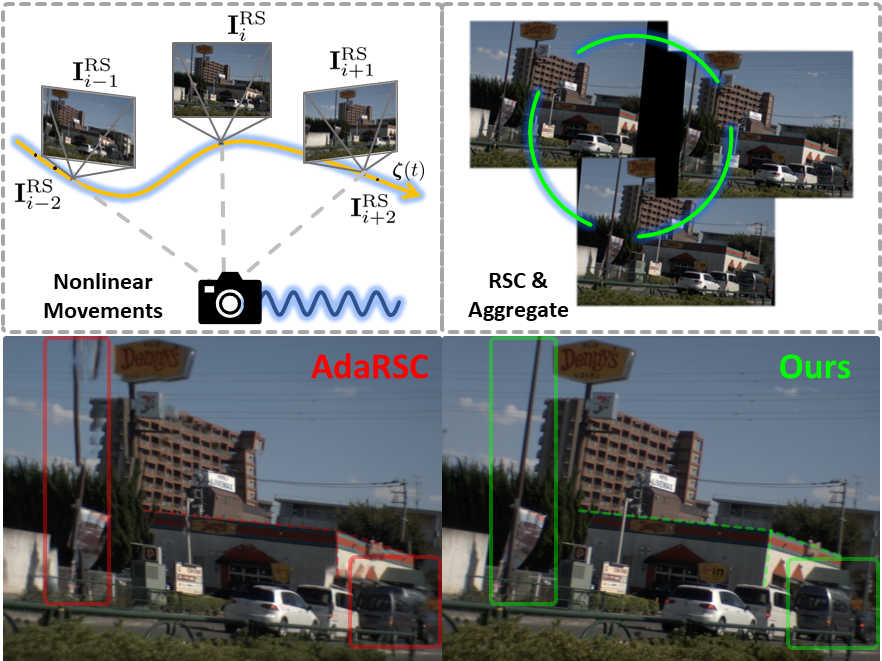}
    \end{center}
    \vspace{-2.ex}
    \caption{Illustration of the challenges in complex nonlinear motion and dynamic scenes with occlusion. The proposed method models the curvilinear trajectory and reconstructs the high-quality occlusion region from adjacent frames of the video stream. In contrast, the state-of-the-art fails to correct the pole and causes significant unaligned artifacts on the bottom right car.}
    \label{fig:head}
    \vspace{-2.5ex}
\end{figure}
The rolling shutter (RS) mechanism widely integrated in consumer video cameras continues to gather photos during the acquisition process, thus effectively increasing sensitivity~\cite{10.1117/12.831203}. The RS cameras donate CMOS sensors and scan the scene sequentially instead of instantly taking a snapshot of the entire scene, like the global shutter (GS) using the CCD sensor~\cite{Meingast2005}. The time slot between consecutive scan lines causes motion artifacts called the RS effect, \textit{e.g.}, wobble and skew, under extreme motion conditions ~\cite{albl2015r6p}. In addition to the detrimental visual artifacts, the RS effect damages numerous 3D vision algorithms based on GS assumptions, such as camera pose estimation~\cite{albl2015r6p, Dai_2016_CVPR}, structure-from-motion~\cite{wu2011visualsfm} and SLAM~\cite{mur2015orb}. Therefore, rolling shutter correction (RSC) is significant in photography and has attracted considerable research attention in the last decades.

Existing works on RS correction are generally categorized into single-frame and multi-frame methods. 
The single-frame methods are based on strict geometric assumptions~\cite{rengarajan2016bows,purkait2017rolling} or simplified camera motion~\cite{rengarajan2017unrolling,lao2018robustified}, thus obtaining unsatisfied results in complex scenes. In comparison, multi-frame methods are more sensible and achieve higher performance~\cite{zhuang2017rolling}. Typically, the correction field between two frames is estimated by a neural block~\cite{Rhemann2011FastCF} to recover the GS frame~\cite{liu2020deep,cao2022learning,fan_CVR_CVPR22}. Nevertheless, existing RSC methods cannot produce satisfying results because of the following limitations:

\textbf{1) Complex higher-order motion:}
Current methods based on cost volume~\cite{liu2020deep,zhong2021towards,sun2018pwc,cao2022learning} face challenges in estimating the accurate correction field since the field cannot be effectively supervised during training~\cite{cao2022learning}. Besides, RSC solutions depending on optical flow use the uniform velocity assumption~\cite{Fan2021InvertingAR, fan_CVR_CVPR22}, ignoring the nonlinear movements. However, the motion in real scenes can be complex and variable, and the inaccuracy of correction fields will accumulate row by row and eventually lead to significant image correction errors \textit{e.g.}, the distorted pole and house corrected by AdaRSC~\cite{cao2022learning} in Fig.~\ref{fig:head}.

\textbf{2) Scene occlusion:} As shown in Fig.~\ref{fig:head}, object edge occlusion around the entity and image border occlusion prevent RSC solutions from better image synthetic quality. Despite the most recent multiple frames method ~\cite{fan_CVR_CVPR22,cao2022learning} trying to compensate for occluded pixels from consecutive frames, the visual performance cannot satisfy the regular application due to the inherent difficulties in aligning and aggregating multiple frames.

To address the challenges, we model the curvilinear trajectory of pixels analytically and propose a geometry-based \textbf{Q}uadratic \textbf{R}olling \textbf{S}hutter (QRS) motion solver, which precisely estimates the high-order correction field of individual pixel based on the forward and backward optical flows. Benefiting from the rigorous modeling of the RS mechanism, the QRS motion solver demonstrates a strong generalization performance across various datasets and handles RS temporal super-resolution tasks~\cite{Fan2021InvertingAR}. Besides, to reconstruct high-quality occlusion frames in extreme scenes, we present a 3D video architecture which effectively \textbf{A}ligns and \textbf{A}ggregates multi-frame context, namely, RSA$^2$-Net. Tab.~\ref{tab:method_compare} exhibits the superiority of the proposed method, and our contributions can be summarized as follows:

\begin{itemize}[leftmargin=10pt]
    \item We analytically model the trajectory in complex nonlinear movements and present a novel geometry-based quadratic rolling shutter motion solver that precisely estimates the high-order correction field of individual pixels.
    \item We propose a self-alignment 3D video architecture for high-quality frame aggregation and synthesis against extreme scene occlusion.
    \item A broad range of evaluations demonstrates the significant superiority and generalization ability of our proposed method over state-of-the-art methods.
\end{itemize}

\begin{table}[t]
    \centering
    \caption{Comparison of the proposed method vs. the state-of-the-art RSC solutions.}
    \label{tab:method_compare}
    \resizebox{0.485\textwidth}{!}{
        \begin{tabular}{ccccccccccccc}
            \toprule
            Method                                         &
            \begin{tabular}[c]{@{}c@{}}DSfM\\ \cite{zhuang2017rolling}\end{tabular}                      &
            \begin{tabular}[c]{@{}c@{}}DSUN\\ \cite{liu2020deep}\end{tabular}                      &
            \begin{tabular}[c]{@{}c@{}}JCD\\ \cite{zhong2021towards}\end{tabular}                      &
            \begin{tabular}[c]{@{}c@{}}SUNet\\ \cite{fan2021sunet}\end{tabular}                      &
            \begin{tabular}[c]{@{}c@{}}RSSR\\ \cite{Fan2021InvertingAR}\end{tabular}                      &
            \begin{tabular}[c]{@{}c@{}}AdaRSC\\ \cite{cao2022learning}\end{tabular}                      &
            \begin{tabular}[c]{@{}c@{}}CVR\\ \cite{fan_CVR_CVPR22}\end{tabular}                     &
            \begin{tabular}[c]{@{}c@{}}Ours\\ \end{tabular}
            \\ \midrule
            \multicolumn{1}{c|}{Dynamic Scene}             &         & $\surd$ & $\surd$ & $\surd$ & $\surd$ & $\surd$ & $\surd$ & $\surd$ \\
            \multicolumn{1}{c|}{Occlusion Scene}           &         &         &         &         &         &         & $\surd$ & $\surd$ \\
            \multicolumn{1}{c|}{High-order Motion}         & $\surd$ &         &         &         &         &         &         & $\surd$ \\
            \multicolumn{1}{c|}{Temporal Super-Resolution} &         &         &         &         & $\surd$ &         & $\surd$ & $\surd$ \\
            \bottomrule
        \end{tabular}
    }
\end{table}

\section{Related Work}
\label{sec:relatework}

\noindent \textbf{Single-frame models.}
To simplify the RSC problem, many works apply different geometric assumptions, such as the straight lines kept straight in ~\cite{rengarajan2016bows}, vanishing direction restraint in~\cite{purkait2017rolling, purkait2017minimal}, and analytical 3D straight line RS projection model in~\cite{lao2018robustified}. Besides, the simplified camera motion is also applied, for instance, the rotation-only model~\cite{rengarajan2017unrolling,purkait2017rolling,lao2018robustified} and Ackerman model~\cite{purkait2017rolling}. Moreover, the first learning-based model is proposed in~\cite{rengarajan2017unrolling} to remove RS from a single distortion image, and Zhuang \textit{et al.}~\cite{zhuang2019learning} further proposed Convolutional Neural Network (CNN)-based method to learning underlying geometry and recover GS image. Recently, Wang \textit{et al.} ~\cite{Wang2022NeuralGS} present an RS removal model with the global reset feature (RSGR). Nevertheless, single-frame models either rely on strong assumptions or depend on inconspicuous features, which causes an unsatisfactory performance.
\begin{figure*}[t]
    \begin{center}
        \includegraphics[width=1\linewidth]{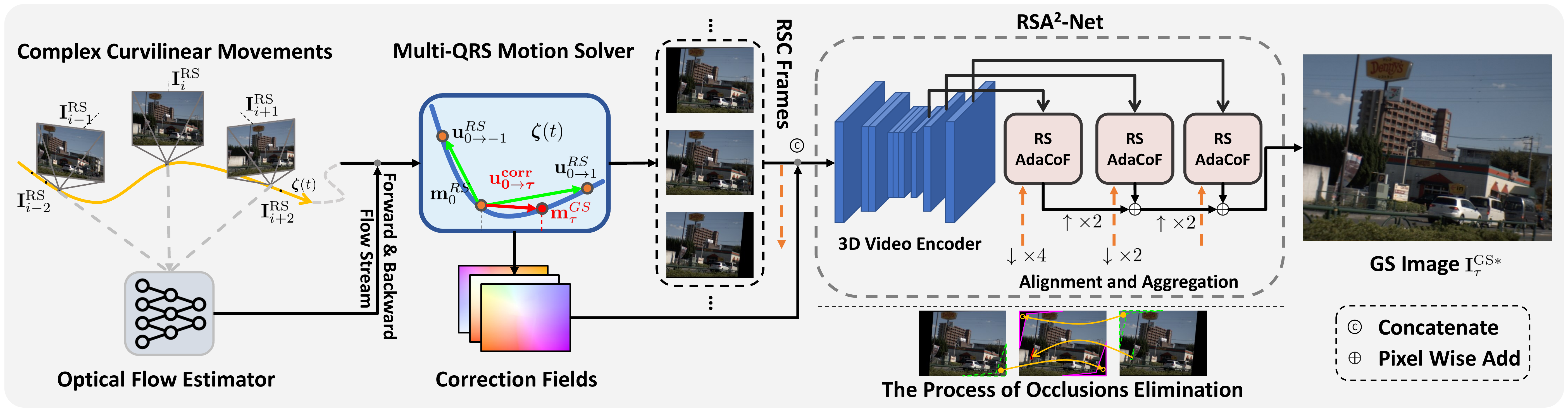}
    \end{center}
    \vspace{-2.ex}
    \caption{Overview of the proposed method. We aim to estimate precise correction fields in nonlinear motion with the QRS motion solver and synthesize high-quality frames against dynamic scenes with extreme occlusion by a self-alignment 3D video architecture RSA$^2$-Net. }
    \label{fig:pipeline}
    \vspace{-2.ex}
\end{figure*}

\noindent \textbf{Multi-frame models.}
Multi-frame methods can be categorized into classical and learning-based models. For the classical multi-frame methods, the works~\cite{grundmann2012calibration, lao2018robust} and~\cite{vasu2018occlusion, zhuang2020image} model the motion between RS frames as a mixture of homography matrices for general unordered RS images and two consecutive frames, respectively. Besides, Zhuang \textit{et al.}~\cite{zhuang2017rolling} develop a solution to estimate relative poses from two consecutive frames and recover GS images based on the differential epipolar constraint under constant velocity and acceleration. However, they either rely on simplified camera motion models or need prior lens calibration.

The learning model based on multiple frames can be divided into motion-field-based and flow-based methods. The former usually computes the cost volumes~\cite{Rhemann2011FastCF}  by a correlation layer to obtain the motion field between two frames. For example, Liu \textit{et al.}~\cite{liu2020deep} designed a deep shutter unrolling network to predict the GS image from two consecutive RS Frames, and Fan \textit{et al.}~\cite{fan2021sunet} present a pyramidal construction to recover the GS image. Besides, Zhong \textit{et al.} ~\cite{zhong2021towards} design an architecture with a deformable attention features fusion module handling both RS distortion and blur. Recently Cao \textit{et al.} ~\cite{cao2022learning} propose an adaptive warping module to warp the RS features into the GS one with multiple displacement fields. Nevertheless, the motion-field-based methods indirectly estimate the correction field, needing adequate supervision\cite{cao2022learning}. As for flow-based methods,~\cite{Fan2021InvertingAR}, and ~\cite{fan_CVR_CVPR22} formulate the RS undistortion flows under the constant velocity assumption indirectly and develop networks to recover RS undistortion from optical flow. However, current flow-based methods ignore the nonlinear movements, thus, fail in complex motion scenes.

\section{Methodology}
\label{sec:methodology}

\subsection{Overview}
Our method aims to estimate precise correction fields in complex nonlinear motion and synthesize high-quality frames against dynamic scenes with occlusion. As shown in Fig.~\ref{fig:pipeline}, the proposed method receives a video stream, typically 5 frames, and precisely corrects the RS images with a geometry-based Multi-QRS motion solver. Then, a 3D video encoder-decoder architecture extracts the preliminary multi-scale features of corrected frames. After that, the hierarchical RSAdaCof modules align and warp the features to produce a final synthetic high-quality GS frame.

\subsection{RSC Formulation}
\label{sec:rsc_formulation}
\begin{figure}[t]
    \vspace{-1ex}
    \begin{center}
        \includegraphics[width=1\linewidth]{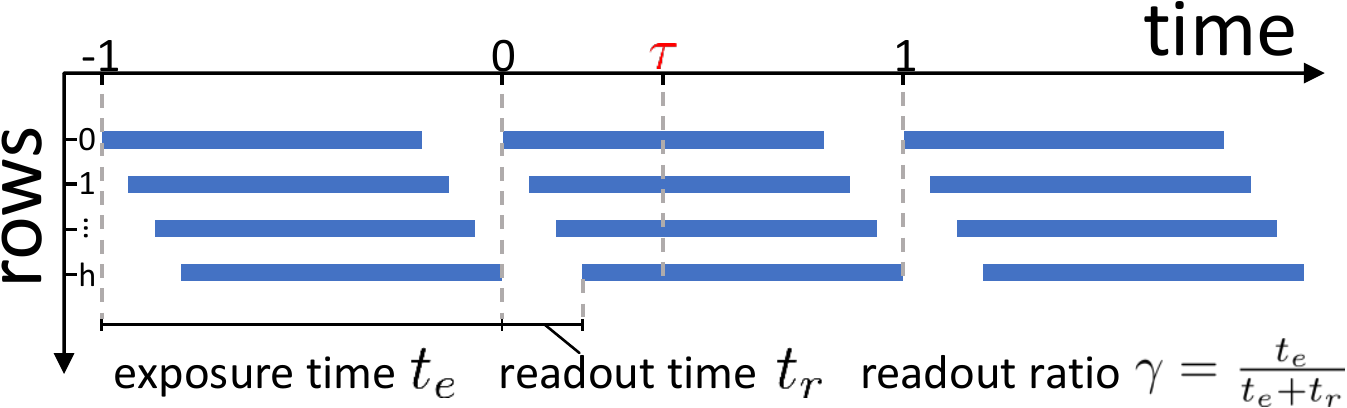}
    \end{center}
    \vspace{-2ex}
    \caption{Illustration of the rolling shutter exposure mechanism among three consecutive frames. The interval time is normalized to $\left[0,1 \right]$, and our goal is to synthesize GS frames at any time $\mathbf{\tau}$ $\in$ $\left[0,1 \right]$.}
    \label{fig:rs_mechanism}
\end{figure}

As the rolling shutter mechanism shown in Fig.~\ref{fig:rs_mechanism}, the CMOS sensor scans the 3D scene sequentially,  and every scanline holds a motion relevant to the row corresponding to the differential timestamp and readout time ratio $\gamma$. Previous work~\cite{liu2020deep} proposes that the GS frame can be recovered by warping the RS features backwards with predicted displacement filed from GS to RS frame $\mathbf{U}_{g \to r}$. However, every single timestamp $\mathbf{\tau}$ between two consecutive frames corresponds to a GS frame, so it is more significant to model the RSC in the entire time series:
\begin{equation}\label{eq:backward}
    \begin{aligned}
        \mathbf{I}^{g(\tau)}(\mathbf{m})=\mathbf{I}^{r}\left(\mathbf{m}+\mathbf{U}_{g(\tau) \rightarrow r}\right),
    \end{aligned}
\end{equation}
where $\mathbf{I}^{g(\tau)}$ denotes the potential GS frame at timestamp $\tau$. $\mathbf{I}^{r}$ is the source RS frame. $\mathbf{m}$ is a specific pixel and $\mathbf{U}_{g(\tau) \rightarrow r}$ is the correction field from GS to RS frame. Relying on many established forward or backward warp techniques (\textit{e.g.}, bilinear interpolation, DFW~\cite{liu2020deep}, and softmax splatting~\cite{niklaus2020softmax}), existing methods achieve pleasing results for a given precise $\mathbf{U}_{g(\tau) \rightarrow r}$. Therefore, calculating the correction field from GS to RS frame is the critical factor of RSC.
\subsection{Quadratic Rolling Shutter Motion Solver}
\begin{figure}[t]
    \begin{center}
        \includegraphics[width=1\linewidth]{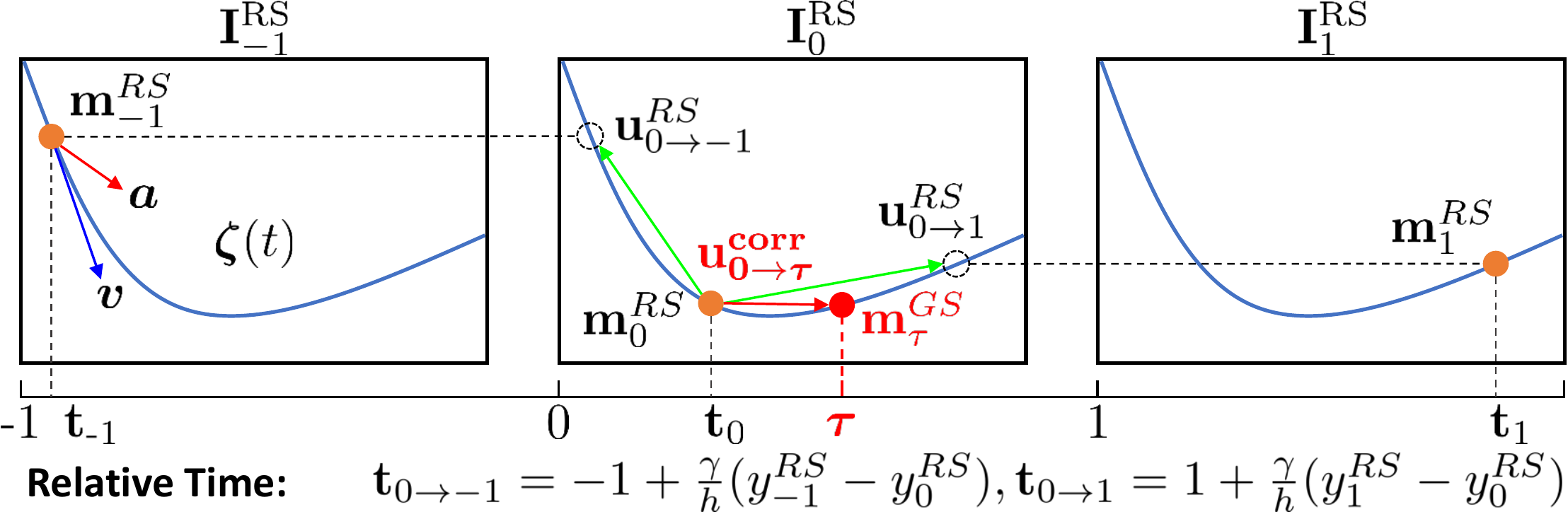}
    \end{center}
    \caption{Illustration of the quadratic motion solver mechanism among three consecutive RS frames. The object moves at variable velocity and has a complex curvilinear trajectory. Note that the object has only 2 Dof in the image plane, and the time interval between frames is very small. Thus, we use a quadratic motion model to formalize the curvilinear trajectory of the pixel. }
    \label{fig:quadratic_motion_solver}
\end{figure}

To address the correction field estimation problem under complex nonlinear motion,  we focus on presenting a precise high-order motion solver for practical variable velocity and dynamic scenes. As the motion scheme shown in Fig~\ref{fig:quadratic_motion_solver}, assuming a 3D point $\mathbf{P}$ is projected to the consecutive three RS image planes $\mathbf{I}_{-1}^\text{RS}$, $\mathbf{I}_{0}^\text{RS}$ and $\mathbf{I}_{1}^\text{RS}$ as image points $\mathbf{m}_{-1}^\text{RS} = [x_{-1}^{\text{RS}}, y_{-1}^{\text{RS}}]^{\top}$, $\mathbf{m}_{0}^\text{RS} = [x_{0}^\text{RS}, y_{0}^\text{RS}]^{\top}$, and $\mathbf{m}_{1}^\text{RS} = [x_{1}^\text{RS}, y_{1}^\text{RS}]^{\top}$, respectively. Despite the fact that the time slot between consecutive frames is extremely temporary, pixels may still follow complex curvilinear movements $\boldsymbol{\zeta}(t)$. According to the derivative definition, we use the second-order Taylor expansion around $t_0$ to formulate as:
\begin{equation}\label{eq:taylor}
    \begin{aligned}
         & \boldsymbol{\zeta}(t) & \approx  \boldsymbol{\zeta}\left(t_{0}\right) + \dot{\boldsymbol{\zeta}}\left(t_{0}\right)\left(t-t_{0}\right) + \frac{1}{2} \ddot{\boldsymbol{\zeta}}\left(t_{0}\right)\left(t-t_{0}\right)^2.
    \end{aligned}
\end{equation}

\noindent Considering the trajectories from $\mathbf{m}_{0}^\text{RS}$ to  $\mathbf{m}_{-1}^\text{RS}$ and $\mathbf{m}_{1}^\text{RS}$, respectively, we obtain:
\begin{equation}
    \small
    \label{eq:linear_equation}
    \begin{aligned}
        \begin{bmatrix}
            (\boldsymbol{\zeta}(t_{-1}) - \boldsymbol{\zeta}\left(t_{0}\right))^{\top} \\
            (\boldsymbol{\zeta}(t_{1}) - \boldsymbol{\zeta}\left(t_{0}\right))^{\top}
        \end{bmatrix} & =
        \begin{bmatrix}
            (t_{-1} - t_{0}) & \frac{(t_{-1} - t_{0})^2}{2} \\
            (t_{1} - t_{0})  & \frac{(t_{1} - t{0})^2}{2}
        \end{bmatrix} \mathbf{M},
        \\
        \mathbf{M}                 & = \begin{bmatrix}
            \dot{\boldsymbol{\zeta}}(t_{0}) & \ddot{\boldsymbol{\zeta}}(t_{0})
        \end{bmatrix}^{\top},
    \end{aligned}
\end{equation}
where $t_{-1}$, $t_{1}$ are the timestamp of $\mathbf{m}_{-1}^\text{RS}$ and $\mathbf{m}_{1}^\text{RS}$. The matrix $\mathbf{M}$ of shape 2 $\times$ 2 measures the quadratic motion of pixels $\mathbf{m}^\text{RS}(t)$ at $t$. Note that the optical flow is the pattern of apparent motion of image objects between two consecutive frames. Thus, the differences of trajectory $\boldsymbol{\zeta}(t)$ are equivalent to the flow vectors $\mathbf{u}_{0 \to -1}^\text{RS}$ from $\mathbf{m}_{0}^\text{RS}$ to $\mathbf{m}_{-1}^\text{RS}$ and $\mathbf{u}_{0 \to 1}^\text{RS}$ from $\mathbf{m}_{0}^\text{RS}$ to $\mathbf{m}_{1}^\text{RS}$.
\begin{equation}
    \small
    \label{eq:flow}
    \begin{aligned}
         & \mathbf{u}_{0 \to -1}^\text{RS} = \boldsymbol{\zeta}(t_{-1}) - \boldsymbol{\zeta}\left(t_{0}\right), \mathbf{u}_{0 \to 1}^\text{RS} = \boldsymbol{\zeta}(t_{1}) - \boldsymbol{\zeta}\left(t_{0}\right).
    \end{aligned}
\end{equation}
According to the scanning mechanism illustrated in Fig.~\ref{fig:rs_mechanism}, the relative time can be expressed as:
\begin{equation}
    \label{eq:relative_time}
    \begin{aligned}
        t_{0 \to -1} & = t_{-1} - t_{0} &  & = -1 + \frac{\gamma}{h}(y_{-1}^\text{RS} - y_0^\text{RS}), \\
        t_{0 \to 1}  & = t_{1} - t_{0}  &  & = 1 + \frac{\gamma}{h}(y_{1}^\text{RS} - y_0^\text{RS}).
    \end{aligned}
\end{equation}
Given the optical flow vectors $\mathbf{u}_{0 \to -1}^\text{RS}$, $\mathbf{u}_{0 \to 1}^\text{RS}$ and the relative times $t_{0 \to -1}$, $t_{0 \to 1}$, we can solve the square system of linear equations in Eq.~(\ref{eq:linear_equation}) and obtain the quadratic motion matrix $\mathbf{M}$ efficiently. Thus, we faithfully compute the correction vector $\mathbf{u}_{0 \to \tau}^\text{corr}$ from $\mathbf{m}_{0}^\text{RS}$ to any timestamp $\boldsymbol{\tau}$ by using:
\begin{equation}
    \label{eq:quadratic_motion_feild}
    \begin{aligned}
        {\mathbf{u}_{0 \to \tau}^\text{corr}}^{\top} & =
        \begin{bmatrix}
            t_{0 \to \tau} & \frac{t_{0 \to \tau}^2}{2}
        \end{bmatrix} \mathbf{M},                                                               \\
        t_{0 \to \tau}                               & = \tau - t_{0} = \tau - \frac{\gamma}{h}y_0^\text{RS}.
    \end{aligned}
\end{equation}

\noindent \textbf{Prime QRS motion solver:} By given three consecutive RS images $\text{I}_\text{-1}^{\text{RS}}$, $\text{I}_\text{0}^{\text{RS}}$ and $\text{I}_\text{2}^{\text{RS}}$, the RS distribution point $\mathbf{m}_{0}^{\text{RS}}$ can be corrected to corresponding GS image measurement $\mathbf{m}_{\tau}^{\text{GS}}$ at any timestamp $\boldsymbol{\tau}$ by following transform:
\begin{equation}\label{equation:quadratic_motion_solver}
    \begin{aligned}
        \mathbf{m}_{\tau}^{\text{GS}} & =   \mathbf{m}_{0}^{\text{RS}} + \mathbf{u}_{0 \to \tau}^\text{corr} \\
                                      & = \mathbf{m}_{0}^{\text{RS}} + \mathbf{M}^{\top}
        \begin{bmatrix}
            t_{0 \to \tau} & \frac{t_{0 \to \tau}^2}{2}
        \end{bmatrix}^{\top}.
    \end{aligned}
\end{equation}
With the precise correction field calculated by QRS motion solver, RS frame $\mathbf{I}_{0}^\text{RS}$ can be effectively restored to corrected frame $\mathbf{I}_{\tau}^\text{RSC}$, by adapting a general bilinear interpolation technique mentioned in Sec.~\ref{sec:rsc_formulation}. As the sample $\mathbf{I}^\text{RSC}_{\tau, 0}$ (green box) shown in Fig.~\ref{fig:QRS}, the proposed prime QRS motion solver is significantly superior to the state-of-the-art~\cite{fan_CVR_CVPR22} according to the visual comparison result without considering image border occlusion.

\subsection{Self-aligning Multi-QRS motion solver}
Although the prime QRS motion solver accurately corrects RS pixels to GS pixels, the object edge occlusion and image border occlusion in Fig.~\ref{fig:head} cannot be restored by the solver. Existing works attempt to compensate these occluded pixels from consecutive two frames~\cite{fan_CVR_CVPR22} by a masked linear aggregation or three~\cite{cao2022learning} frames via a self-attention-based warping module. Nevertheless, except for accurate corrected vector field estimation, they suffer from inherent difficulties in aligning and aggregating multiple frames. 

\begin{figure}[t]
    \begin{center}
        \includegraphics[width=1\linewidth]{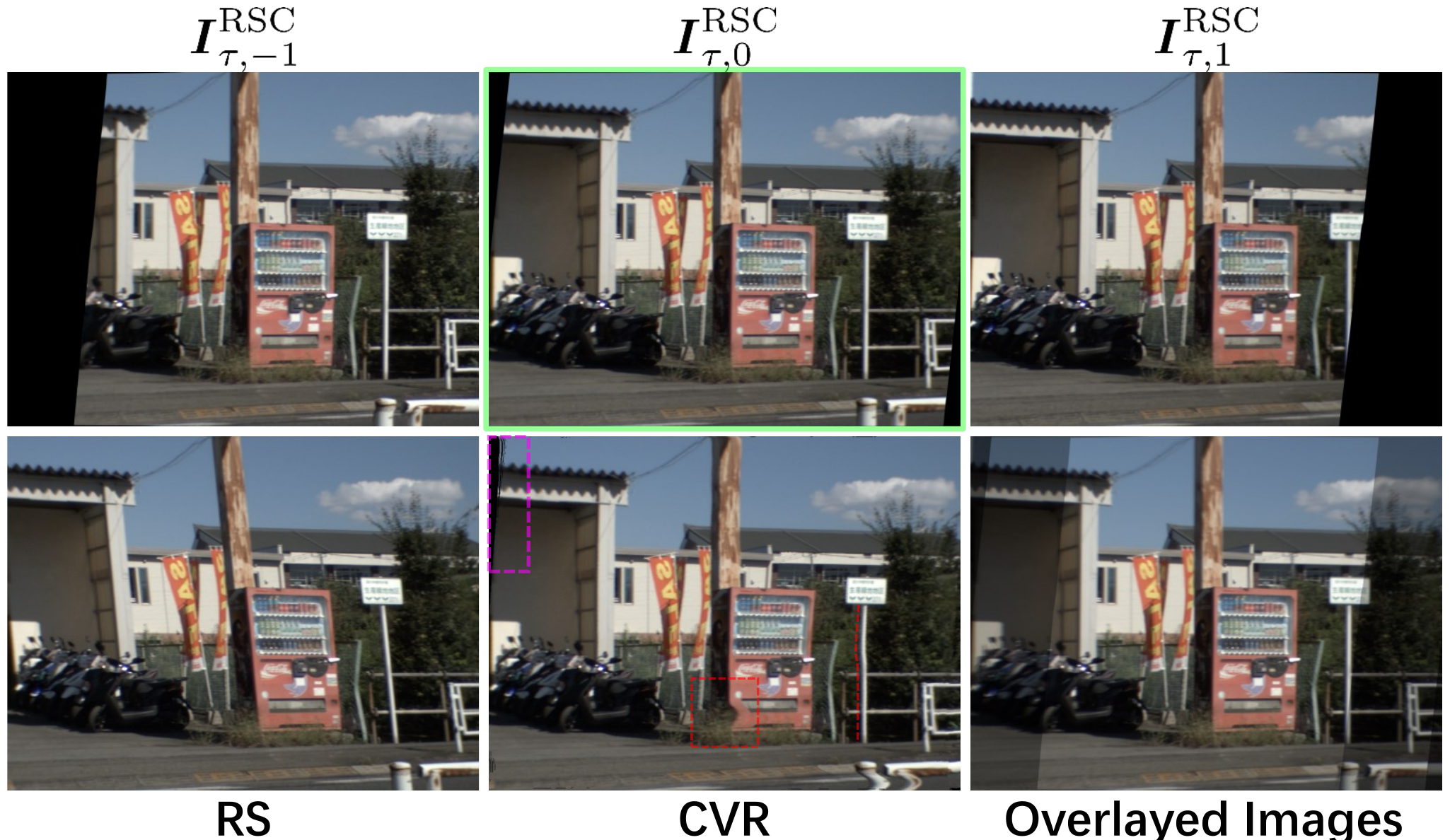}
    \end{center}
    \vspace{-2.ex}
    \caption{An example of the proposed Multi-QRS motion solver correcting from 5 frames. The overlapping image is directly average from three RSC image pixels.}
    \label{fig:QRS}
\end{figure}

\noindent \textbf{Multi-QRS motion solver:} In order to address the above problem, we design a self-aligning algorithm by extending the QRS motion solver to a video sequence. Considering a consecutive RS video sequence $\boldsymbol{I}_{\left\{0,1, \cdots,N - 1\right\}}$ consisting of ${N}$ frames, we aim to correct all the pixels $\mathbf{m}_{i}^{RS}$ from the neighbour frames $\mathbf{I}_i^\text{RS}$, $\mathbf{I}_i^\text{RS} \in \boldsymbol{I}_{\left\{0,1, \cdots,N - 1\right\}}$ to time $\tau$, usually in the intermediate, for occlusion elimination and obtain $\mathbf{I}_{\tau,i}^\text{RSC}$. Replacing $-1$, $0$ and $1$ with $i-1$, $i$ and $i+1$ at Eq.~(\ref{eq:taylor})(\ref{eq:linear_equation})(\ref{eq:flow})(\ref{eq:relative_time})(\ref{eq:quadratic_motion_feild}) in order we obtain:
\begin{equation}
    \label{eq:multi_quadratic_motion_solver}
    \begin{aligned}
         & \mathbf{m}_{\tau,i}^{\text{GS}} &  & =   \mathbf{m}_{i}^{\text{RS}} + \mathbf{M}^{\top}_{i} \begin{bmatrix}
            t_{i \to \tau} & \frac{t_{i \to \tau}^2}{2}
        \end{bmatrix}^{\top}, \\
         & t_{i \to \tau}                  &  & = \tau - t_{i} = \tau - \frac{\gamma}{h}y_i^\text{RS}, i \in \left [  1, N-2 \right ] ,
    \end{aligned}
\end{equation}
where $t_{i \to \tau}$ is the relative time from time $t_{i}$ to $\tau$. The matrix $\mathbf{M}_i$ denotes the quadratic motion of $\mathbf{m}_{i}^{\text{RS}}$. And the $\mathbf{m}_{\tau,i}^{\text{GS}}$ is the corrected GS pixel from $\mathbf{m}_{i}^{\text{RS}}$ = $[x_{i}^\text{RS}, y_{i}^\text{RS}]^{\top}$. The Multi-QRS motion solver corrects pixels on different RS frames and naturally aligns pixels to time $\tau$ by Eq.~(\ref{eq:multi_quadratic_motion_solver}), which accurately models the timeline of the consecutive frames. As shown in Fig.~\ref{fig:QRS}, RS frames $\mathbf{I}_{-1}^\text{RS}$, $\mathbf{I}_{0}^\text{RS}$ and $\mathbf{I}_{1}^\text{RS}$ are warped to time $\tau$ to obtain $\mathbf{I}_{\tau,-1}^\text{RSC}$, $\mathbf{I}_{\tau,0}^\text{RSC}$ and $\mathbf{I}_{\tau,1}^\text{RSC}$. By contributing a simple average fusion, we achieve a seamless and complete composite overlayed image.

\subsection{3D Video RSA$^2$-Net}
\label{sec:RSANet}
Given the corrected frames $\mathbf{I}_{\tau,1}^\text{RSC}$, $\mathbf{I}_{\tau,2}^\text{RSC}$, $\cdots$, $\mathbf{I}_{\tau, N-2}^\text{RSC}$, we aim to eliminate extreme occlusions and synthesize a high-quality GS frames $\mathbf{I}_{\tau}^{\text{GS}\ast}$ through the multi-frame fusion function:
\begin{equation}
    \label{eq:fusion_function}
    \begin{aligned}
        \mathbf{I}_{\tau}^{\text{GS}\ast} = {f}(\mathbf{I}_{\tau,1}^{\text{RSC}}, \mathbf{I}_{\tau,2}^{\text{RSC}},\cdots,\mathbf{I}_{\tau, N-2}^{\text{RSC}}; \boldsymbol{\theta}),
    \end{aligned}
\end{equation}
where $\boldsymbol{\theta}$ is the parameters of function ${f}$. We developed a 3D RS video encoder-decoder architecture RSA$^2$-Net, which receives a video RSC stream and correction fields, to seek the fusion function ${f}$ for extreme occlusions elimination and high-quality GS frames synthetic.

\noindent \textbf{Model Architecture:}
As shown in Fig.~\ref{fig:pipeline}, the RSA$^2$-Net comprises a 3D video encoder (which is a 3D-transformer in our model), the sequentially arranged Decode Layer, and iterative RSAdaCof modules. The encoder receives the sequence input of shape ${B \times T \times C \times H \times W}$, which is concatenated from consecutive frames and the corresponding correction field $\mathbf{u}_{i \to \tau}^\text{corr}$ obtained by the Multi-QRS motion solver. Then 3D video encoder encodes the input in four stages to obtain the multi-scale features, where $B$, $C$, $T$, $H$, and $W$ respectively denote the batch size, channel, time, height, and width dimensions. The Encoded features are decoded with three times $2\times$ upsampling, and then hierarchically warped by the RSAdaCof model at scale $l$, where $l \in \left\{ 1,2,3 \right\}$, to produce a final synthetic high-quality GS Frame $\mathbf{I}_{\tau}^{\text{GS}\ast}$.

\begin{figure}[t]
    \begin{center}
        \includegraphics[width=1\linewidth]{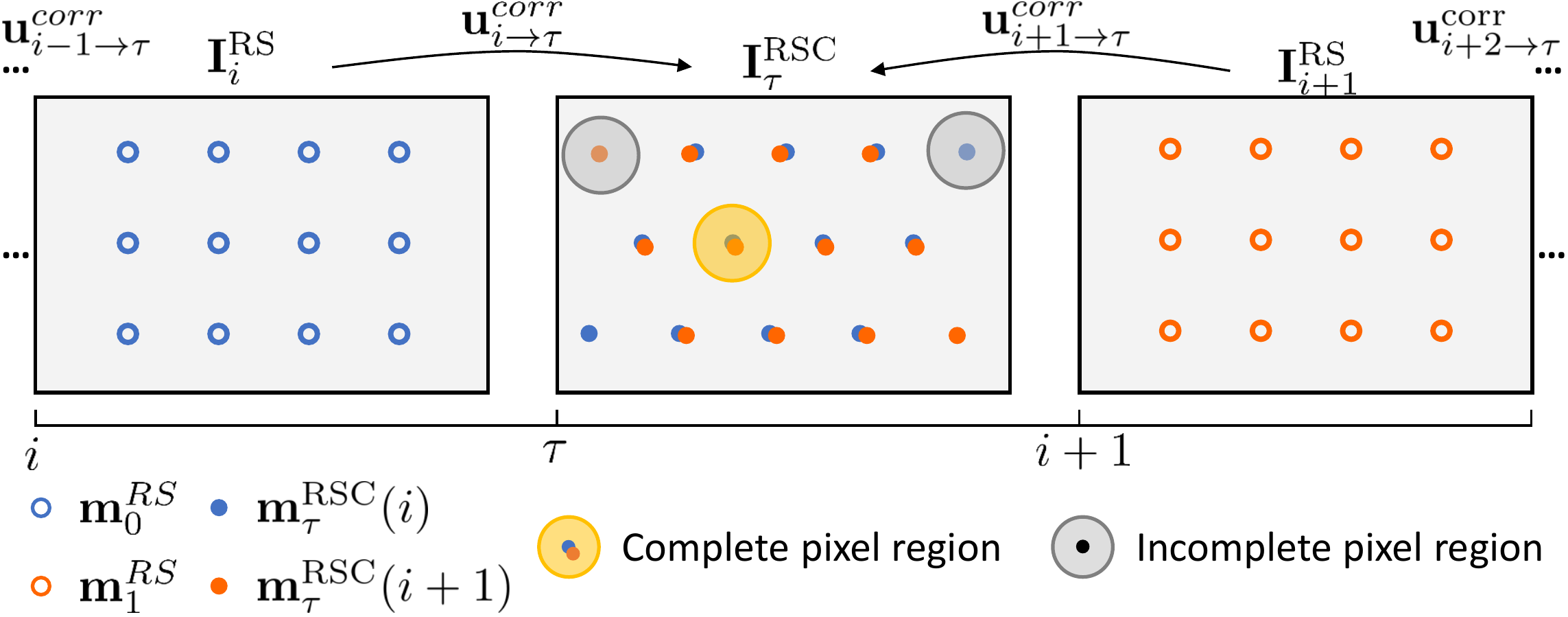}
    \end{center}
    \vspace{-2.ex}
    \caption{Illustration of the multi-frames alignment and fusion scheme. Slight offsets exist between the aligned pixels in extreme motion, and incomplete regions cover only pixels from the individual frame.}
    \label{fig:fusion_scheme}
\end{figure}

\noindent \textbf{RSAdaCof Warping:}
Multi-frames are aligned to a specific time $\boldsymbol{\tau}$, as shown in Fig.~\ref{fig:fusion_scheme}. However, slight horizontal or vertical offsets might exist between the RSC pixel and GS pixel in extreme motion, which is not fully modeled by the QRS motion solver. Inspired by the AdaCof~\cite{lee2020adacof} and ~\cite{shi2022video}, we adopt three CNN-based offset Nets to obtain the per-pixel deformable kernels, which represent offsets $\boldsymbol{\alpha}_{i}^{l}(k,x,y)$ and $\boldsymbol{\beta}_{i}^{l}(k,x,y)$, and the weights ${W}_{i}^{l}(k,x,y)$, for frame $\mathbf{I}_{\tau,i,l}^{RSC}$ at scale $l$. Thus, the intermediate pixel $O_{i}^{l}(x, y)$ at position $[x, y]^{\top}$ of frame $\mathbf{I}_{\tau,i,l}^{RSC}$ is:
\begin{equation}
    \label{eq:intermediate_pixel}
    \resizebox{0.4125\textwidth}{!}{
        $O_{i}^{l}(x, y)=\sum_{k=1}^{K} W_{i}^{l}(k, x, y) \mathbf{I}_{\tau,i,l}^{RSC}\left(x+\alpha_{i}^{l}(k, x, y), y+\beta_{i}^{l}(k, x, y)\right)$
    },
\end{equation}
where $K$ is the number of sampling locations of each kernel. To align and aggregate the incomplete pixel region and the complete pixel region shown in Fig.~\ref{fig:fusion_scheme}, we use a CNN with softmax layer to predict the mask $M_i^l$ corresponding to frame $\mathbf{I}_{\tau,i,l}^{RSC}$. However, considering the RS scanline mechanism, different pixels hold individual relative positions, and the quadratic model degrades gradually with increasing temporal distance (imagine using RS Frames at $t$ to recover GS frames at infinity time $\tau$). Thus, we propose to dilute the degradation using the nearest neighbor principle, which uses completely independent differential temporal distances (time grid ${tg}^l$) to weigh the intensities of different pixels:
\begin{equation}
    \label{eq:syn_frame}
    \resizebox{0.85\linewidth}{!}{
        \begin{math}
            \begin{aligned}
                 & O^{l}=\sum_{i} (1 - \frac{tg^l}{S_{tg}^l}) M_{i}^{l} \cdot O_{i}^{l}, \quad tg^l = \left | \tau + \left \lfloor \frac{N - 2}{2} \right \rfloor - i - \frac{\gamma}{h^l}y^l \right |, \\
                 & S_{tg}^l = \sum_{i}^{N-2} tg^{l}, i \in \left [  1, N-2 \right ],
            \end{aligned}
        \end{math}
    }
\end{equation}
where $h^l$ and $y^l$ denote the image height and differential time grid at scale $l$. Note that the BS-RSC~\cite{cao2022learning} dataset retains the $\mathbf{\gamma} = 0.45$, meaning there are 55\% blank rows between two consecutive frames. And this ultra-long time distance is a huge challenge for network training, so in general, we fix $\mathbf{\gamma} = 1$ in RSAdaCof for rapid convergence.

\subsection{Loss Functions}
Following~\cite{zhong2021towards,cao2022learning}, we use the Charbonnier loss $\mathcal{L}_{c}$ and perceptual loss $\mathcal{L}_{p}$ to ensure the synthetic visual quality, and the MSE $\mathcal{L}_{mse}$ to avoid the extreme pixels. A simple but effective multi-loss balancing technique is adapted, \textit{i.e.} scaling them to the same scale as the first. Thus total loss can be formulated as follows:
\begin{equation}
    \label{eq:loss}
    \begin{aligned}
        \mathcal{L} & =\mathcal{L}_{c} + \lambda_{p} \mathcal{L}_{p} +  \lambda_{mse}\mathcal{L}_{mse},                                                                                               \\
        \lambda_{p} & = \frac{\left | \mathcal{L}_{c} \right | }{\left  | \mathcal{L}_{p} \right | }, \lambda_{mse} = \frac{\left | \mathcal{L}_{c} \right | }{\left  | \mathcal{L}_{mse} \right | }.
    \end{aligned}
\end{equation}

\section{Experiment}
\subsection{Experimental Setup}
\noindent \textbf{Datasets.}
We evaluate the proposed RSC method on the Carla-RS, Fastec-RS~\cite{liu2020deep}, BS-RSC~\cite{cao2022learning}, and ACC datasets. The synthetic Carla-RS contains general 6-DoF camera motions, and Fastec-RS holds multiple challenging dynamic scenes. Besides, the lately released real-world dataset BS-RSC~\cite{cao2022learning} consists of nonlinear movements. Moreover, we derived the ACC dataset, which comprises more challenging variable movements, by excluding frames with constant general motions from the BS-RSC dataset (details can be found in the supplementary material). In addition, we provide more visual comparisons on other datasets in the supplementary material, including GPark~\cite{jia2012probabilistic}, Seq77~\cite{Kim}, 3GS and House~\cite{forssen2010rectifying}.

\noindent \textbf{Evaluation Metrics and Comparison methods.}
We use the PSNR, SSIM, and LPIPS to evaluate the quantitative results of RSC methods. In Sec.~\ref{sec:metric}, we report the Carla-RS with the mask (CRM), Carla-RS without the mask (CR), Fastec-RS (FR), BS-RSC (BR), and ACC. Note that to force our method to fully integrate information among adjacent frames, the GS frames corresponding to the middle row are donated as supervision, and $\tau$ is set to 0.5 $\gamma$. We compare the proposed method with state-of-the-art RSC methods, including geometry-based methods \textbf{DSfM}~\cite{zhuang2017rolling}, \textbf{DiffHomo}~\cite{zhuang2020image}, and learning-based methods \textbf{DSUN}~\cite{liu2020deep}, \textbf{SUNet}~\cite{fan2021sunet}, \textbf{JCD}~\cite{zhong2021towards}, \textbf{RSSR}~\cite{Fan2021InvertingAR}, \textbf{VideoRS}~\cite{naor2022combining}, \textbf{CVR}~\cite{fan_CVR_CVPR22} and \textbf{AdaRSC}~\cite{cao2022learning}. The models of most methods are unavailable on BS-RSC as it's recently released, so we collected the models of DSUN and JCD provided by Cao \textit{et al.}~\cite{cao2022learning} and trained CVR~\cite{fan_CVR_CVPR22} model on BS-RSC dataset using official released code. (Experiments of temporal super-resolution can be found in the supplementary material).

 \noindent \textbf{Implementation Details.} We use RAFT~\cite{teed2020raft} and GMA~\cite{jiang2021learning} of OpenMMLab Optical Flow Toolbox~\cite{2021mmflow} to predict optical flow from the $N=5$ consecutive RS frames, followed by  Multi-QRS motion solver to predict the correction fields and obtain the latent occluded three RSC frames. Besides, we set the image readout ratio $\gamma$ for Carla-RS, Fastec-RS, BS-RSC, and ACC datasets to 1.0, 1.0, 0.45, and 0.45, respectively, based on the intrinsic parameters of each dataset. The model is trained for 80 epochs using the Adam optimizer with learning rate $1e^{-4}$, batch size 4, and StepLR scheduler with gamma 0.1 and step size 25.

\subsection{Quantitative Analysis}
\label{sec:metric}

\begin{table}[t]
    \centering
    \caption{Quantitative comparison against the state-of-the-art RSC methods on the Carla-RS dataset.}
    \label{tab:metric_carla}
    \resizebox{0.48\textwidth}{!}{
        \begin{tabular}{lcccccc}
            \toprule
                                            & \multicolumn{2}{c}{PSNR $\uparrow$ (dB)} &                   & \multicolumn{1}{c}{SSIM $\uparrow$(dB)} &                   & \multicolumn{1}{c}{LPIPS $\downarrow$}                      \\
            \cline{2-3} \cline{5-5} \cline{7-7}
            \multirow{-2}{*}{Method}        & CRM                                      & CR                &                                         & CR                &                                        & CR                 \\
            \midrule
            DSfM~\cite{zhuang2017rolling}   & 24.20                                    & 21.28             &                                         & 0.775             &                                        & 0.1322             \\
            DiffHomo~\cite{zhuang2020image} & 19.60                                    & 18.94             &                                         & 0.606             &                                        & 0.1798             \\
            \midrule
            DSUN~\cite{liu2020deep}         & 26.90                                    & 26.46             &                                         & 0.807             &                                        & 0.0703             \\
            SUNet~\cite{fan2021sunet}       & 29.28                                    & 29.18             &                                         & 0.850             &                                        & 0.0658             \\
            RSSR~\cite{Fan2021InvertingAR}  & 30.17                                    & 24.78             &                                         & 0.867             &                                        & 0.0695             \\
            VideoRS~\cite{naor2022combining}& 31.84                                    & 31.43             &                                         & 0.919             &                                        & --------            \\
            \rowcolor[HTML]{EFEFEF}
            CVR~\cite{fan_CVR_CVPR22}       & \underline{32.02}                        & \underline{31.74} &                                         & \underline{0.929} &                                        & \underline{0.0368} \\
            \midrule
            \rowcolor[HTML]{EFEFEF}
            Ours                            & {\textbf{37.00}}                         & {\textbf{32.01}}  &                                         & {\textbf{0.933}}  &                                        & {\textbf{0.0253}}  \\
            \bottomrule
        \end{tabular}
    }
    \vspace{-1ex}
\end{table}

\noindent \textbf{Results on Carla-RS and Fastec-RS.}
The results reported in Tab.~\ref{tab:metric_carla} and Tab.~\ref{tab:metric_fastec} show that the proposed method outperforms the other eight RSC methods by large margins on both datasets, especially 37.00 (PSNR) compared to 32.02 on CRM and 0.0814 (LPIPS) against 0.1107 on FR achieved by CVR~\cite{fan_CVR_CVPR22}. These superior performances significantly demonstrate the effectiveness of our model on general 6 Dof and highly dynamic scenes with occlusion.

\begin{table}[t!]
    \centering
    \caption{Quantitative comparison against the state-of-the-art RSC methods on the Fastec-RS dataset.}
    \label{tab:metric_fastec}
    \resizebox{0.48\textwidth}{!}{
        \begin{tabular}{lccc}
            \toprule
            Method                          & PSNR $\uparrow$ (dB) & SSIM $\uparrow$(dB) & LPIPS $\downarrow$  \\
            \midrule
            DSfM~\cite{zhuang2017rolling}   & 20.14                & 0.701               & 0.1789              \\
            DiffHomo~\cite{zhuang2020image} & 18.68                & 0.609               & 0.2229              \\
            \midrule
            DSUN~\cite{liu2020deep}         & 26.52                & 0.792               & 0.1222              \\
            SUNet~\cite{fan2021sunet}       & 28.34                & 0.837               & 0.1205              \\
            JCD~\cite{zhong2021towards}     & 24.84                & 0.778               & 0.1070               \\
            RSSR~\cite{Fan2021InvertingAR}  & 21.23                & 0.776               & 0.1659              \\
            VideoRS~\cite{naor2022combining}& 28.57                & 0.844               & --------         \\
            \rowcolor[HTML]{EFEFEF}
            AdaRSC~\cite{cao2022learning}   & 28.56                & \underline{0.855}  & 0.0793              \\
            \rowcolor[HTML]{EFEFEF}
            CVR~\cite{fan_CVR_CVPR22}       & \underline{28.72}   & 0.847               & \underline{0.1107} \\
            \midrule
            \rowcolor[HTML]{EFEFEF}
            Ours                            & {\textbf{29.49}}     & {\textbf{0.872}}    & {\textbf{0.0814}}   \\
            \bottomrule
        \end{tabular}
    }
\end{table}

\noindent \textbf{Results on BS-RSC and ACC.}
The quantitative comparison on the real-world curvilinear movement BS-RSC and ACC datasets is shown in Tab.~\ref{tab:metric_bsrsc_acc}. The proposed method dramatically surpasses the existing five RSC methods with a PSNR score of 33.50 and an SSIM score of 0.946 compared to 29.17 and 0.896 achieved by the second-best entry AdaRSC on the BS-RSC dataset. Note that all the other methods produce a significant drop when testing on ACC, but our method remains stable. It demonstrates the effectiveness and robustness of our method in handling complex motion and acceleration trajectories.

\begin{table}[t]
    \centering
    \caption{Quantitative comparison against the state-of-the-art RSC methods on the BS-RSC and ACC datasets.}
    \label{tab:metric_bsrsc_acc}
    \resizebox{0.48\textwidth}{!}{
        \begin{tabular}{lcccccccc}
            \toprule
            \multirow{2}{*}{Method}       & \multicolumn{2}{c}{PSNR $\uparrow$ (dB)} &                   & \multicolumn{2}{c}{SSIM$\uparrow$(dB)} &                   & \multicolumn{2}{c}{LPIPS$\downarrow$}                                              \\ \cline{2-3} \cline{5-6} \cline{8-9}
                                          & BR                                       & ACC               &                                        & BR                & ACC                                   &  & BR                 & ACC                \\
            \midrule
            DSfM~\cite{zhuang2017rolling} & 19.80                                    & 15.74             &                                        & 0.698             & 0.551                                 &  & 0.2437             & 0.2544             \\
            DSUN~\cite{liu2020deep}       & 23.60                                    & 22.39             &                                        & 0.808             & 0.780                                 &  & 0.1035             & 0.1145             \\
            JCD~\cite{zhong2021towards}   & 24.86                                    & 23.73             &                                        & 0.820             & 0.808                                 &  & 0.1897             & 0.1961             \\
            CVR~\cite{fan_CVR_CVPR22}     & 24.58                                    & 24.07             &                                        & 0.823             & 0.816                                 &  & 0.0795             & 0.0822             \\
            \rowcolor[HTML]{EFEFEF}
            AdaRSC~\cite{cao2022learning} & \underline{29.17}                        & \underline{28.73} &                                        & \underline{0.896} & \underline{0.892}                     &  & \underline{0.0617} & \underline{0.0637} \\
            \midrule
            \rowcolor[HTML]{EFEFEF}
            Ours                          & { \textbf{33.50}}                        & { \textbf{33.36}} &                                        & { \textbf{0.946}} & { \textbf{0.945}}                     &  & { \textbf{0.0299}} & { \textbf{0.0303}} \\
            \bottomrule
        \end{tabular}
    }
    \vspace{-1ex}
\end{table}

\begin{figure*}[t]
    \begin{center}
        \includegraphics[width=1\linewidth]{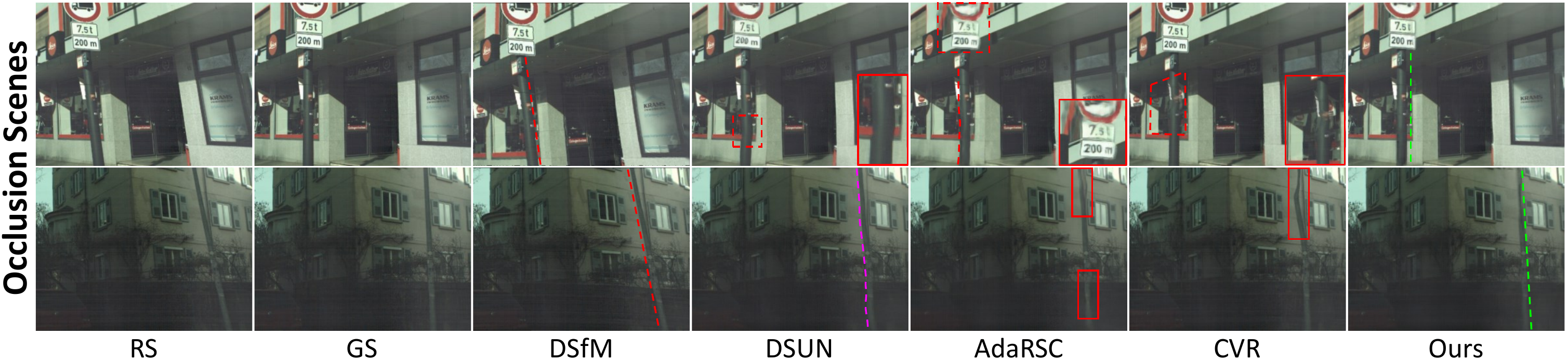}
    \end{center}
    \vspace{-2ex}
    \caption{Visual comparison against the state-of-the-art RSC methods in dynamic and occlusion scenes on Fastec-RS datasets.}
    \label{fig:visual_compare_fastec}
\end{figure*}

\begin{figure*}[t]
    \begin{center}
        \includegraphics[width=1\linewidth]{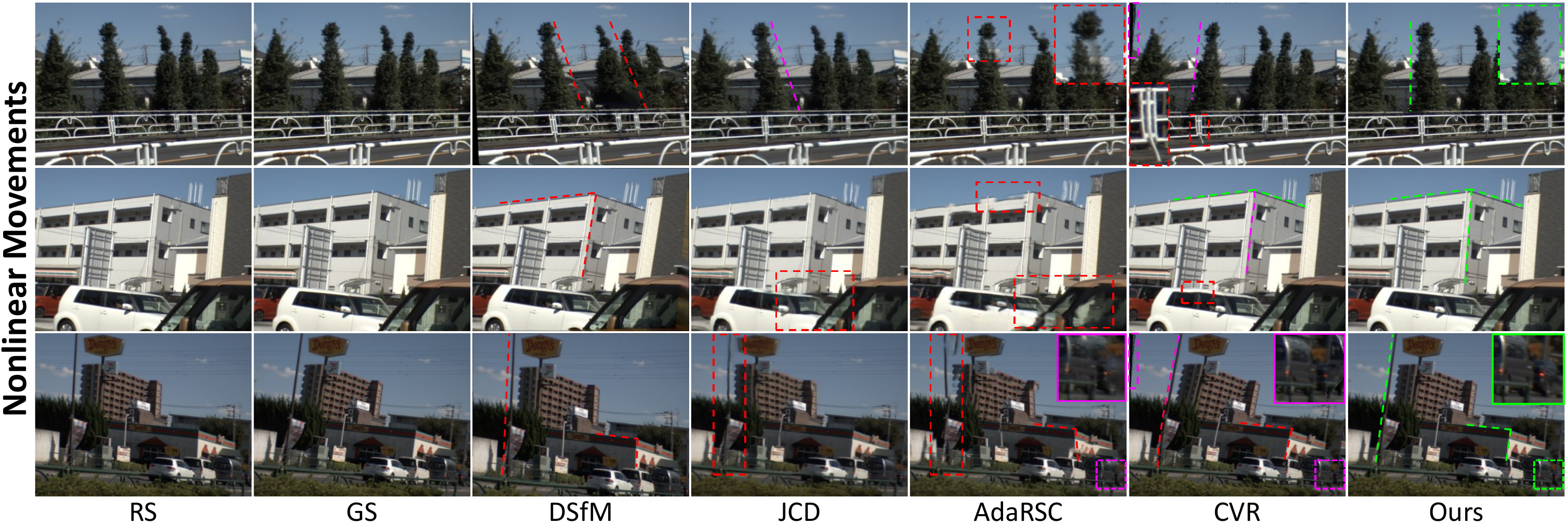}
    \end{center}
    \vspace{-2.ex}
    \caption{Visual comparison against the state-of-the-art RSC methods in nonlinear motion scenes of BS-RSC dataset.}
    \label{fig:visual_compare_bsrsc}
    \vspace{-1ex}
\end{figure*}

\subsection{Visual Comparisons}
\noindent \textbf{Dynamic and occlusion scenes:}
The comparisons of dynamic and highly occluded scenes are reported in Fig.~\ref{fig:visual_compare_fastec} containing various moving objects with different depths and motions. Only the proposed method rectifies the poles back to the right position naturally, while the others either fail in correction (DSfM and DSUN) or produce artifacts on the occlusion areas (AdaRSC and CVR), although CVR was designed to handle occlusion.

\noindent \textbf{Nonlinear Movements:} As the results shown in Fig.~\ref{fig:visual_compare_bsrsc}, the proposed method precisely restored the GS images in curvilinear motion. Similarly, all existing RSC solutions fail in such a nonlinear scene because of the incorrectly estimated correction fields. For example, the tree on the first row and pole on the third row are inaccurately corrected by DSfM, JCD and CVR, and AdaRSC produces noticeable artifacts. Besides, the sota methods CVR and AdaRSC cause significant unaligned shadow when handling occlusion in nonlinear scenes, \textit{e.g.}, the bottom right car of the third row (more results can be found in the supplementary material).

\subsection{Generalization Ability}
\begin{figure}[t]
    \begin{center}
        \includegraphics[width=1\linewidth]{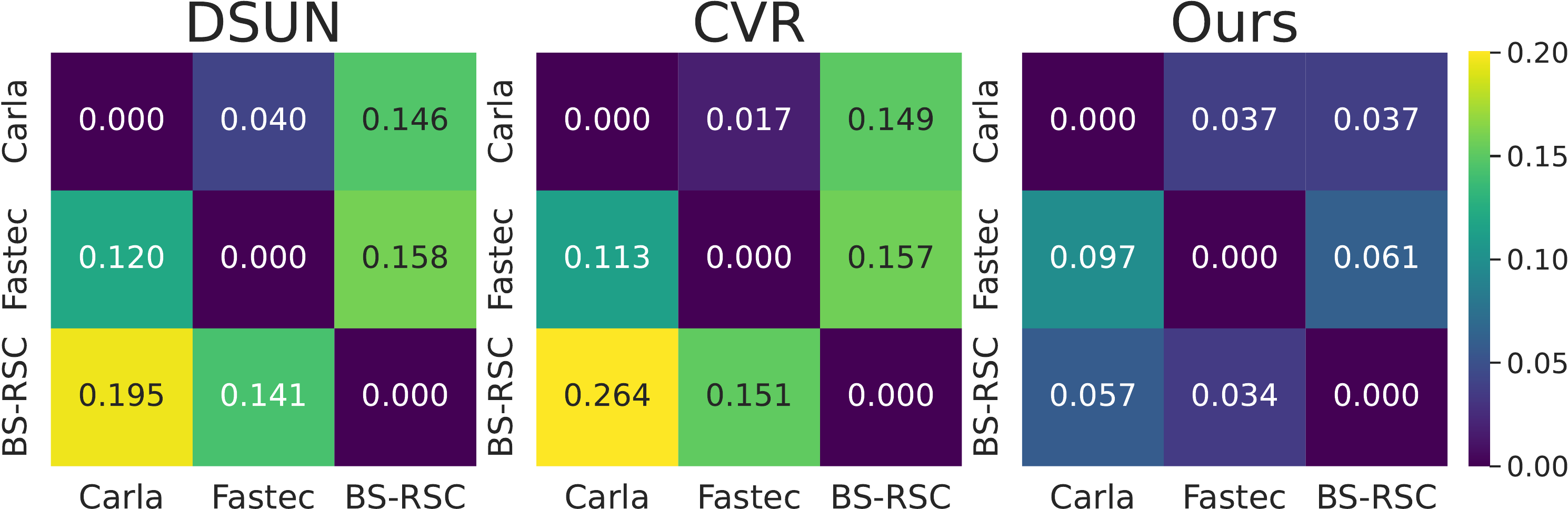}
    \end{center}
    \vspace{-1.ex}
    \caption{Generalization capability comparisons on \textbf{SSIM} of our method with existing RSC algorithms DSUN~\cite{liu2020deep} and CVR~\cite{fan_CVR_CVPR22} across Carla-RS, Fastec-RS and BS-RSC datasets.}
    \label{fig:generalize}
    \vspace{-2ex}
\end{figure}

To validate the generalization capability of the proposed method, we performed cross-tests on three datasets and donated a relative decline rate ${rde}(i,k)$ for evaluation:
\begin{equation}
    \label{eq:rde}
    \begin{aligned}
        {rde}_{i,k} = 1 - \frac{{score}_{i,k}}{{score}_{k,k}}, \\
    \end{aligned}
\end{equation}
where ${rde}_{i,k}$ and ${score}_{i,k}$ respectively denote the $rde$ and metric score (PSNR, SSIM, and LPIPS), which is trained on dataset $i$ and tested on dataset $k$. The confusion matrixes in Fig.~\ref{fig:generalize} show that DSUN and CVR cannot accommodate numerous datasets, especially across the BS-RSC dataset. This is because the intrinsic readout ratio $\gamma$ significantly changes from 1 (Carla-RS or Fasstec-RS) to 0.45 (BS-RSC). In contrast, benefiting from the QRS motion solver, which analytical modeling of the RS mechanism, the proposed method demonstrates strong generalization performance with ${rde}$ less than 0.1, despite the massive bias among three datasets.

\begin{table*}[h!]
    \centering
    \caption{Ablation study results for RSAdaCof, correction field $\mathbf{u}$, 3D video encoder, number of input frames (NF).}
    \label{tab:ablation}
    \vspace{-1.ex}
    \resizebox{0.98\textwidth}{!}{
        \begin{threeparttable}
            \begin{tabular}{lcccccccccccc}
                \toprule
                \multirow{2}{*}{Settings} & \multicolumn{4}{c}{PSNR$\uparrow$ (dB)} &                & \multicolumn{3}{c}{SSIM$\uparrow$ (dB)} &                & \multicolumn{3}{c}{LPIPS$\downarrow$}                                                                                                             \\ \cline{2-5} \cline{7-9} \cline{11-13}
                                          & CRM                                     & CR             & FR                                      & BR             &                                       & CR             & FR             & BR             &  & CR              & FR              & BR              \\
                \midrule
                W/o correction field $\mathbf{u}$
                                          & 36.00                                   & 30.92          & 28.72                                   & 32.36          &                                       & 0.922          & 0.872          & 0.938          &  & 0.0304          & 0.0878          & 0.0331          \\
                W/o RSAdaCof              & 35.90                                   & 30.84          & 28.70                                   & 32.75          &                                       & 0.923          & 0.863          & 0.943          &  & 0.0266          & 0.0881          & 0.0314          \\
                QRS motion solver + 3DUNet       & 36.15                                   & 31.06          & 28.51                                   & 33.41          &                                       & 0.929          & 0.865          & 0.946          &  & 0.0267          & 0.0827          & 0.0306          \\
                \rowcolor[HTML]{EFEFEF}
                Full model       & \textbf{37.00}                          & \textbf{32.01} & \textbf{29.49}                          & \textbf{33.50} &                                       & \textbf{0.933} & \textbf{0.872} & \textbf{0.946} &  & \textbf{0.0253} & \textbf{0.0814} & \textbf{0.0299} \\
                \midrule
                3NF input                 & 35.64                                   & 29.81          & 28.18                                   & 31.92          &                                       & 0.919          & 0.8530         & 0.942          &  & 0.0313          & 0.0912          & 0.0320          \\
                4NF input                 & 35.95                                   & 30.98          & 28.26                                   & 32.40          &                                       & 0.925          & 0.8538         & 0.944          &  & 0.0282          & 0.0901          & 0.0303          \\
                \rowcolor[HTML]{EFEFEF}
                5NF input       & \textbf{37.00}                          & \textbf{32.01} & \textbf{29.49}                          & \textbf{33.50} &                                       & \textbf{0.933} & \textbf{0.872} & \textbf{0.946} &  & \textbf{0.0253} & \textbf{0.0814} & \textbf{0.0299} \\
                \bottomrule
            \end{tabular}
        \end{threeparttable}
    }
\end{table*}

\section{Ablation Study}
\label{sec:AB}
\begin{figure}[t!]
    \vspace{-1.ex}
    \begin{center}
        \includegraphics[width=1.\linewidth]{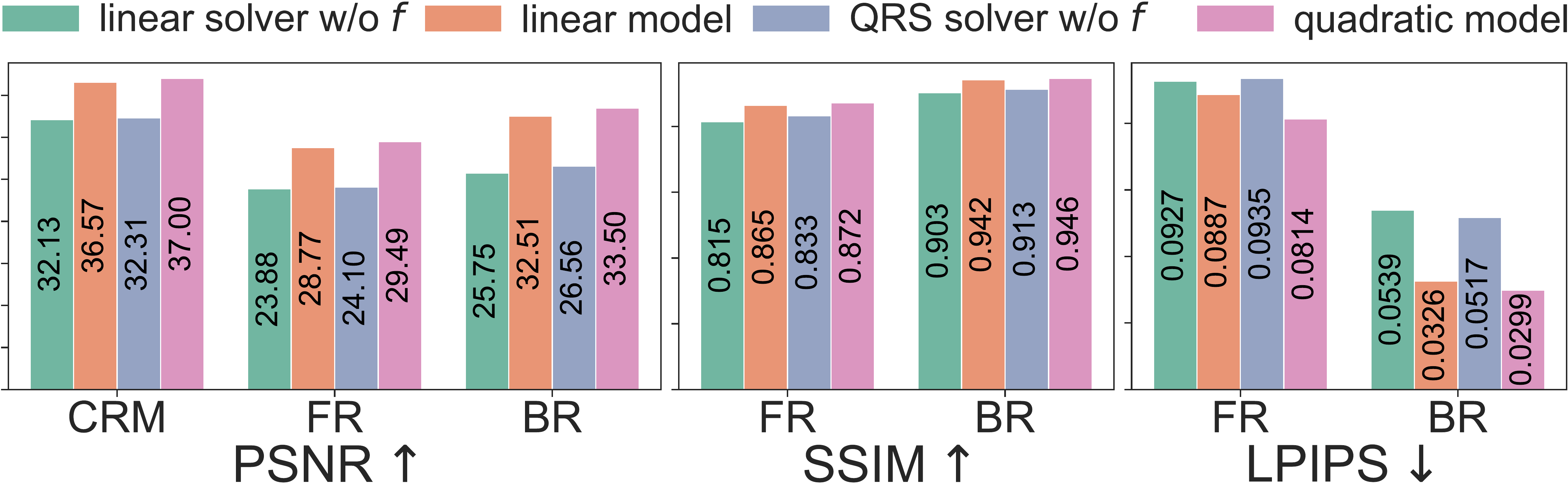}
    \end{center}
    \vspace{-2ex}
    \caption{Ablation study of Linear and Quadratic Models.}
    \label{fig:AB_qudratic_linear}
    \vspace{-2ex}
\end{figure}

\noindent \textbf{Linear Model $\boldsymbol{vs}$ Quadratic Model.}
This experiment compares the proposed quadratic and the linear model (the first-order form of QRS motion solver) with and without RSA$^2$-Net ${f}$, respectively. Noting the solver generates RSC frames containing image occlusion, thus, it is more meaningful to focus on PSNR scores with Mask. As Fig.~\ref{fig:AB_qudratic_linear} shows, they all achieve high metric scores in the Carla-RS dataset. And the quadratic model outperforms the linear model, especially on the Fastec-RS and BS-RSC datasets containing complex variable velocity scenes, illustrating the validity of higher-order models under nonlinear motion.

\begin{figure}[t!]
    \vspace{-1.ex}
    \begin{center}
        \includegraphics[width=1.\linewidth]{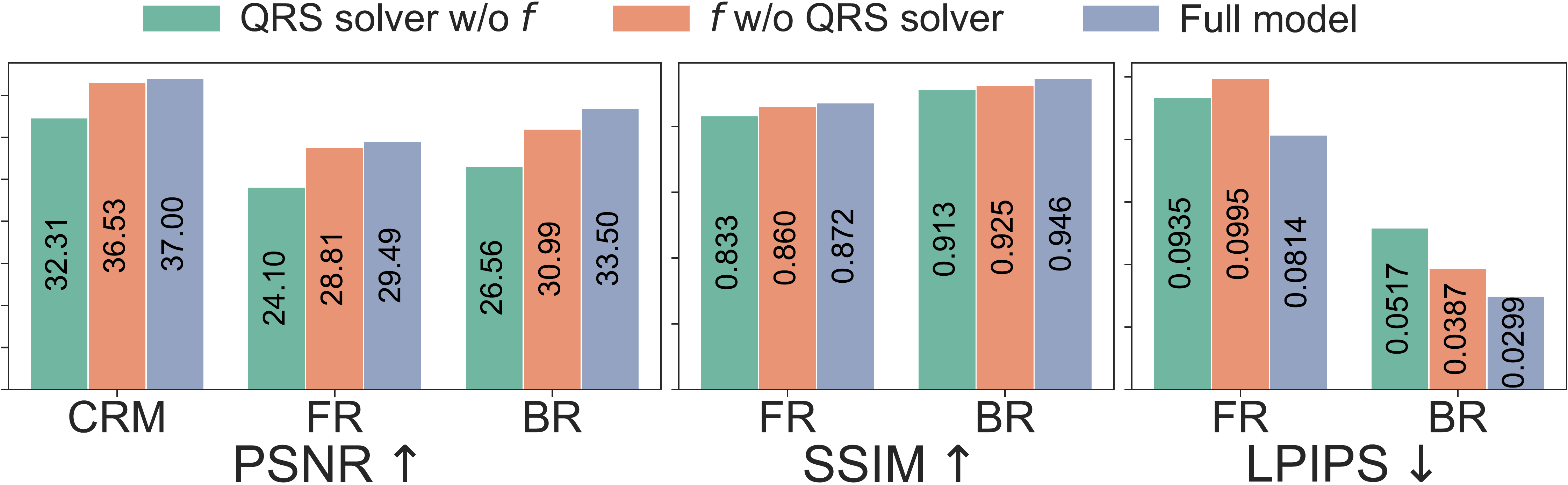}
    \end{center}
    \vspace{-2ex}
    \caption{Ablation study of QRS solver and RSA$^2$-Net.}
    \label{fig:AB_qrs_net}
    \vspace{-2.ex}
\end{figure}

\noindent \textbf{QRS motion solver and RSA$^2$-Net.}
We remove the RSA$^2$-Net ${f}$ or QRS motion solver from the pipeline to validate the capability of correction field estimation and occlusion elimination. As shown in Fig.~\ref{fig:AB_qrs_net}, the QRS motion solver achieves ultra-high PSNR (32.31, CM), SSIM (0.913, BR), and lower LPIPS (0.0517, BR), and even outperform 32.02, 0.896 and 0.0617 achieved by the state-of-the-art. Besides, the single RSA$^2$-Net performs fairly high metric scores on all three datasets, displaying the strong ability of content aggregation. In addition, the performance is significantly improved once the QRS motion solver and RSA$^2$-Net are combined, especially on the Fasstec-RS and BS-RSC datasets containing complex motion.

\noindent \textbf{RSAdaCof and correction field.}
This experiment eliminated the RSAdaCof and correction field $\mathbf{u}$ to validate their effectiveness separately. The results reported in Tab.~\ref{tab:ablation} show significant performance degradation after removing module RSAdaCof due to the lack of alignment and aggregation capabilities. Similarly, our method achieves a lower score without correction field $\mathbf{u}$ because it helps the model map RSC pixels back to RS planes.

\noindent \textbf{3D-Transformer $\boldsymbol{vs}$ 3D-UNet.}
In this experiment, we further replace the 3D-Transformer in 3D Video RSA$^2$-Net with a quite simple 3D-UNet to determine the effectiveness of the proposed QRS motion solver and RSAdaCof. The results in Tab.~\ref{tab:ablation} show that even the simple 3D-UNet with QRS motion solver significantly outperforms the state-of-the-art methods and achieves similar scores with the Full model. It illustrates that our method does not rely on the ability of the Transformer architecture and confirms the effectiveness of the QRS motion solver and RSAdaCof module.

\noindent \textbf{Number of Input Frames.}
We argue that exquisite alignment and fusion techniques based on the video sequence can solve the object edge and image border occlusion. According to the performance in Tab.~\ref{tab:ablation}, the evaluation score increases with the number of input frames, especially in extreme scenes. However, we did not test more than five frames due to computational limitations.

\section{Conclusion and Limitation}
\label{sec:limitation_conclusion}
This paper proposes a geometry-based quadratic rolling shutter motion solver and the 3D video stream-based structure RSA$^2$-Net for RSC in complex nonlinear scenes with occlusion. A broad range of evaluations demonstrates the significant superiority of our proposed method over state-of-the-art methods. However, the proposed method requires dense matching between multiple consecutive frames, which can be expensive in some application scenes. This limitation is a common constraint among the most state-of-the-art RSC solutions~\cite{zhuang2017rolling,Fan2021InvertingAR,fan_CVR_CVPR22}. In future work, we aim to extend the QRS motion solver to sparse keypoint correction, serving for the 3D vision algorithm instead of the dense visual correction, which allows GS SfM/SLAM solutions to handle RS input in real time.

\noindent \textbf{Acknowledgements.} This research is supported by the Shanghai AI Laboratory and the National Natural Science Foundation of China under
Grant 62106183 and 62102145.

{\small
    \bibliographystyle{ieee_fullname}
    \bibliography{egbib}
}

\end{document}